\documentclass[runningheads]{llncs}
\usepackage{graphicx}

\usepackage{tikz}
\usepackage{comment}
\usepackage{amsmath,amssymb} 
\usepackage{color}

\usepackage[accsupp]{axessibility}  
\usepackage{threeparttable}
\usepackage{multirow}
\usepackage{graphicx}

\begin{document}
\pagestyle{headings}
\mainmatter
\def\ECCVSubNumber{10}  


\title{UnconFuse: Avatar Reconstruction from Unconstrained Images} 

\titlerunning{ECCV-22 submission ID \ECCVSubNumber} 
\authorrunning{ECCV-22 submission ID \ECCVSubNumber} 
\author{Anonymous ECCV submission}
\institute{Paper ID \ECCVSubNumber}

\titlerunning{UnconFuse: Avatar Reconstruction from Unconstrained Images}
%
\author{Han Huang\inst{1}\index{van Author, First E.} \and
Liliang Chen\inst{1,2} \and
Xihao Wang\inst{1,3}}
\authorrunning{H. Huang et al.}
%
\institute{OPPO Research Institute, Beijing, China\\ 
\email{huanghan@oppo.com}\\\and
University of Pittsburgh, Pittsburgh, USA
\and
Xidian University, Xian, China\\
}
\maketitle

\begin{abstract}
The report proposes an effective solution about 3D human body reconstruction from multiple unconstrained frames for ECCV 2022 WCPA Challenge: From Face, Body and Fashion to 3D Virtual avatars I (track1: Multi-View Based 3D Human Body Reconstruction). We reproduce the reconstruction method presented in MVP-Human~\cite{zhu2022mvp} as our baseline, and make some improvements for the particularity of this challenge. We finally achieve the score 0.93 on the official testing set, getting the 1st place on the leaderboard.


\end{abstract}

\section{Method}
The purpose of this challenge is to explore various methods to reconstruct a high-quality T-pose 3D human  in the canonical space from multiple unconstrained frames where the character can have totally different actions. Many existing works \cite{saito2019pifu,Saito_2020_CVPR,zheng2021pamir,xiu2022icon,Huang_2020_CVPR} have proved that pixel-aligned image features are significant to recover details of the body surface, and spatially-aligned local features queried from multi-view images can greatly improve the robustness of the reconstructions since more regions of the body can be observed. These methods use well-calibrated camera parameters to ensure the accurate inter-view alignment and 3D-to-2D alignment, while the setup of multiple unconstrained frames makes it much difficult to achieve precision alignments. Recent years, tracking based methods are explored to reconstruct human with single-view video or unconstrained multiple frames \cite{zheng2021pamir,li2021posefusion,DoubleFusion,zhu2022mvp,crosshuman}. To our knowledge, the research purpose of MVP-Human~\cite{zhu2022mvp} is most related to our challenge, aiming to reconstructing an avatar consisting of geometry and skinning weights~\cite{kavan2007skinning} from multiple unconstrained images. We reproduce the methods proposed in MVP-Human and make some modifications. In this section, we will deliver a introduction about the details of our pipeline and experiments settings through which we have achieved a fairly high accuracy on the challenging MVP-Human dataset~\cite{zhu2022mvp}.   

In general, we query pixel-aligned image features from several images and fuse them through parametric body motion. Then, occupancy field of the canonical space are predicted from these fused features. Like MVP-Human, normals from SMPL models are also introduced to improve the reconstruction quality. Considering the requirements of this challenge, we focus more on geometry reconstruction quality in canonical space and only use one backbone for both occupancy and skinning weights prediction.
Our experiments demonstrate that our data preprocessing, data augmentation, model architecture design, feature aggregation strategy and online hard example mining(OHEM)~\cite{shrivastava2016training} all contribute to higher score.

\subsection{Data Preprocessing}
Data preprocessing is the basis of the entire task, which strongly affects the reconstruction accuracy. We perform the following processes for training and testing data:

\textbf{Human detection and image crop.}
The size of the raw images is 1280 $\times$ 720 and the character in the image is lying down. We rotate the images and detect the human bounding box with off-the-shelf method HumanDet~\cite{HumanDet}. To make images more friendly to encoder networks, we crop the them to 512 $\times$ 512 and ensure the character roughly on the center of the image.

\textbf{Segmentation.}
We extract the foreground mask with the resolution 512 $\times$ 512 through PGN~\cite{gong2018instance} .

\textbf{SMPL Fitting.}
We apply the state-of-the-art method PyMAF~\cite{zhang2021pymaf} to fit SMPL parameters for each image. Ideally, one character with different actions or view-point should share the same SMPL shape, but the shape parameters extracted frame by frame are slightly different. To regularize the shape consistency of the same character, we take a joint optimization strategy for SMPL fitting, achieving only one set of shape parameters for each character. The fitting error at the beginning and the ending of iteration of the joint optimization process are respectively shown in Fig. 1 and Fig. 2, we can see
that the fitting error have been definitely minimized when joint optimization
finished.

\textbf{Sampling.}
Like the training strategy proposed in PIFu~\cite{saito2019pifu}, we sample points in the 3D space and around the iso-surface. For each training iteration, we sample 14756 points close to the ground-truth body surface and 1628 points randomly distributed in a 3D bounding box containing the ground-truth body.

\textbf{Normals.}
In order to better utilize the prior knowledge of the SMPL model, we calculate the point-wise normals of the SMPL, and use K-Nearest Neighbor (KNN) algorithm to obtain the normal information of all sampling points.

\textbf{Dataset split.}
Considering that the challenge doesn't provide the ground-truth for testing set, we manually split the training data to be training set and validation set. Specifically, we randomly select 190 subjects from all 200 subjects for training, and the rest 10 subjects for validation. 

\textbf{Data Augmentation.}
In order to achieve a model with better generalizability, we attempt to enhance the diversity of the training data by image augmentation which simulates various texture, material or lighting conditions at the image level. We randomly jitter the brightness, hue, saturation, and contrast where brightness factor, saturation and contrast is all chosen as 0.2, hue is set to be 0.05.

\begin{figure}[h]
\centering
\includegraphics[height=3.2cm]{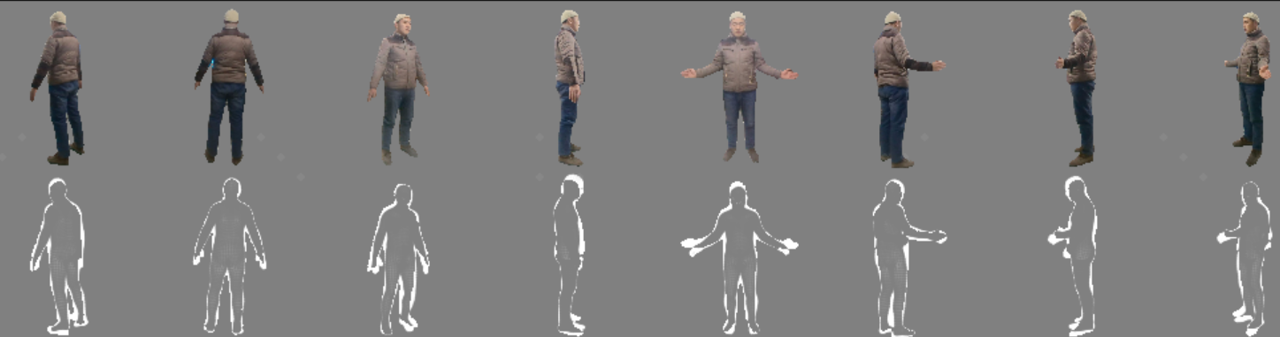}
\caption{Part of results at the beginning of the iteration. On the Bottom row, the white contour represents the fitting error and the thinner the contour width, the smaller the error is.\\}
\label{fig:img1}
\end{figure}

\begin{figure}[h]
\centering
\includegraphics[height=3.15cm]{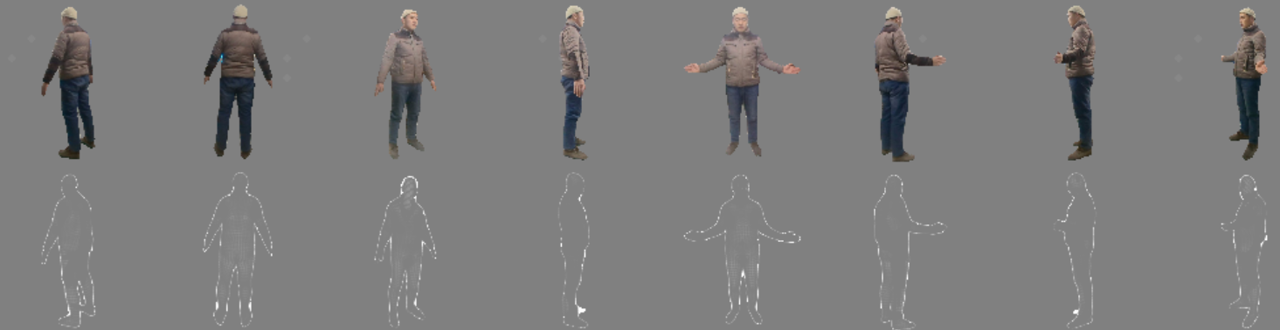}
\caption{Part of results at the end of the iteration. On the Bottom row, the white contour represents the fitting error and the thinner the contour width, the smaller the error is.}
\label{fig:img2}
\end{figure}

\subsection{Model Design and Configuration}
The architecture of our method is shown in \textbf{Fig. 3}. As mentioned above, this challenge aims to reconstruct high quality T-pose human in the canonical space with detailed garment from multiple unconstrained images. Referring to the framework of MVP-Human~\cite{zhu2022mvp}, voxels or sampling points in canonical spaces are warped with SMPL motion to align with posed images. Then, occupancy and skinning weights are predicted from pixel-aligned features through MLPs. Finally, the explicit mesh will be extracted through Marching Cubes~\cite{lorensen1987marching}. The loss functions for training our model is quite similar to the ones used in PIFu and MVP-Human. Since this challenge care more about the geometry in canonical space, we don't assign a specialized backbone for skinning weight prediction. Instead of the stacked hourglass network used in MVP-Human, we introduce HRNet as image encoder that shows better reconstruction accuracy and also faster runtime according to MonoPort~\cite{li2020monoport,li2020monoportRTL}.

In addition, we perform the position encoding~\cite{mildenhall2020nerf} to the 3D coordinates of queried points in the canonical space and concatenate them with corresponding local image features and noramls extracted from SMPL body surface.

In our baseline method, we simply apply average pooling to fuse features from different frames. In order to weigh the contribution of image features from different view-point to a 3D point, we introduce the self-attention mechanism proposed in DeepMultiCap~\cite{zheng2021deepmulticap}  for feature aggregation. Our experiments proved that such feature fusion strategy leads to better performance in this task.

\begin{figure}[t]
\centering
\includegraphics[scale=0.34,trim={1cm 7cm 7cm 0cm},clip]{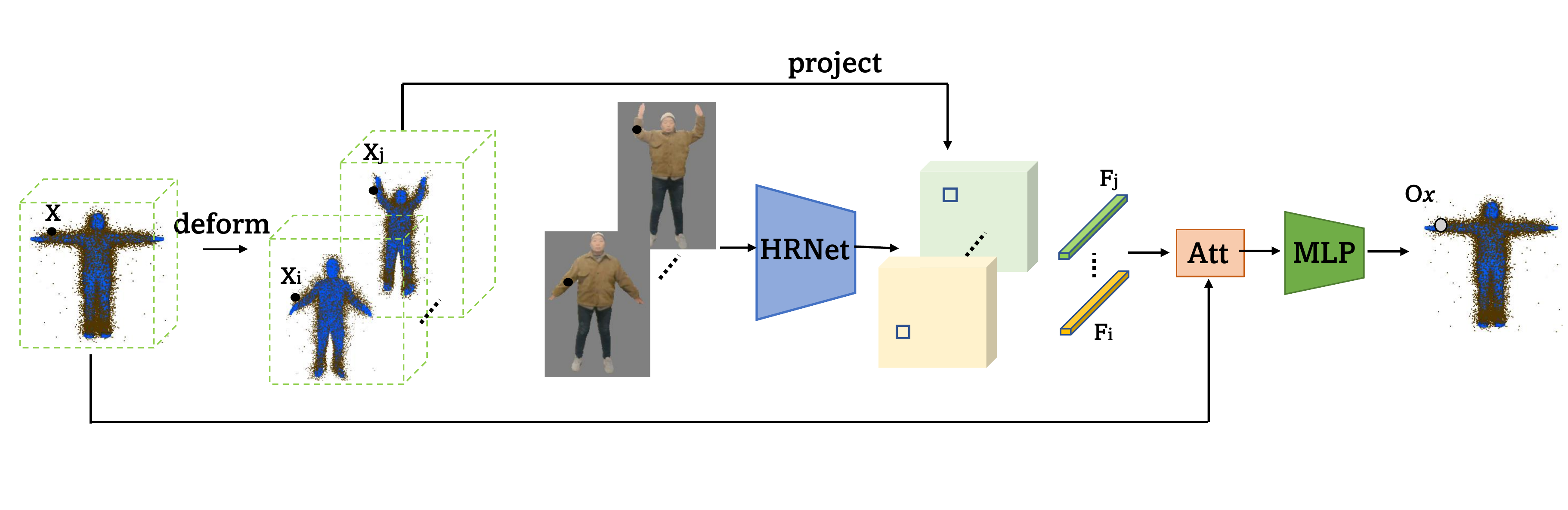}
\caption{The overview of our model architecture. Points in the canonical space are warped with SMPL motion to align with unconstrained multi-frame images. Then, local image features are fused by self-attention. Finally, the occupancy or skinning weights are predicted from fused features through MLPs.  \\}
\label{fig:model}
\end{figure}

\section{Results}
As shown in \textbf{Table 1}, after all strategies mentioned above are used, our model finally achieve score (chamfer distance with ground-truth mesh) 0.93 on the testing set. 


\begin{table*}[]
\caption{Quantitative results of our methods}
\begin{threeparttable}
\setlength{\tabcolsep}{7.5mm}{
\resizebox{\textwidth}{!}{%
\begin{tabular}{l|c}
\hline
\multirow{2}{*}{Our methods with different settings}                               & \multirow{2}{*}{Chamfer$\downarrow$} \\
                                                &                          \\ \hline
our baseline                                    & 1.13                     \\
baseline + PE                                   & 1.09                     \\
baseline + PE + Att                             & 0.96                     \\
baseline + PE + Att + DA                        & 0.95                     \\
Our final model (baseline + PE + Att + DA + HM) & 0.93                     \\ \hline
\end{tabular}%


}}

   \begin{tablenotes}
     \item PE: position encoding for points coordinates, Att: self-attention for multi-frame feature fusion, DA: data augmentation, HM: online hard case mining.
   \end{tablenotes}
\end{threeparttable}
\end{table*}

\newpage

\section{Discussion}
We appreciate the challenge for providing this valuable academic exchange opportunities. The novel and interesting setup contributes to the innovation of this specific research area. We are glad to explore the avatar reconstruction methods and lead the race. Though our quantitative results look good and get 1st place on the leaderboard, the meshes generated by our method are still very smooth without high-fidelity details. We will further improve our methods for better reconstruction quality under limited data.

\clearpage
%
%
\bibliographystyle{splncs04}
\bibliography{egbib.bib}
\end{document}